\documentclass{ijuc}
\usepackage[]{graphicx}

\begin{document}

% MA - added quotation marks to title and changed my affiliation to reflect 
%      internal shake-up

\title{``Going Back to our Roots": Second Generation Biocomputing}

\author{Jon Timmis \inst{1}\email{jtimmis@cs.york.ac.uk}, Martyn Amos
\inst{2}\email{M.R.Amos@exeter.ac.uk}, Wolfgang Banzhaf \inst{3}\email{banzhaf@cs.mun.ca}, Andy Tyrrell \inst{4}\email{amt@ohm.york.ac.uk}}

\institute{Department of Electronics and Department of Computer Science, University of York, YO10 5DD, UK
\and School of Engineering, Computer Science and Mathematics, University of Exeter, Exeter EX4 4QF, UK
\and Department of Computer Science, Memorial University of Newfoundland, St. John's, NL, A1B 3X5, Canada.
\and Department of Electronics, University of York, YO10 5DD, UK}

\def\received{Received 16th December; }

\maketitle
\begin{abstract}

%Bio-inspired systems have plundered the natural world for inspiration, making
%some significant progress over the years, creating systems that are robust,
%adaptable and develop solutions to problems that humans may never have
%identified.  However, this paper argues, it is now time for the bio-inspired
%community, to adopt a more interdisciplinary approach, and rather than
%plunder and run, work closely with the biologists to exploit the richness in
%biological systems, that paradigms fail to capture.  To facilitate our
%arguments, we examine four areas of bioinspired computing: genetic
%programming and artificial immune systems, evolvable hardware and natural
%genetic engineering.

% MA - modified abstract - please feel free to ignore!

Researchers in the field of biocomputing have, for many years, successfully
``harvested and exploited" the natural world for inspiration in developing systems that are
robust, adaptable and capable of generating novel and even ``creative"
solutions to human-defined problems. However, in this position paper we argue
that the time has now come for a reassessment of how we exploit biology to 
generate new computational systems. Previous solutions (the ``first
generation" of biocomputing techniques), whilst reasonably
effective, are crude analogues of actual biological systems. We believe that
a new, inherently inter-disciplinary approach is needed for the development
of the emerging ``second generation" of bio-inspired methods. This new 
{\it modus operandi} will require much closer interaction between the 
engineering and life sciences communities, as well as a bidirectional flow
of concepts, applications and expertise.  We support our argument by examining,
in this new light, three existing areas of biocomputing (genetic programming,
artificial immune systems and evolvable hardware), as well as an emerging
area (natural genetic engineering) which may provide useful pointers as to
the way forward.

\end{abstract}


% MA - added inline attribution to place quotation in historial context

\section{Introduction}
\begin{quote}
Natural organisms are, as a rule, much more complicated and subtle, and
therefore much less well understood in detail, than are artificial automata.
Nevertheless, some regularities which we observe in the organisation of the
former may be quite instructive in our thinking and planning of the latter -- John von Neumann, 1948 
\cite{von-neumann48}.
\end{quote}

Even just after the 2nd world war, scientists were already thinking about the
conceptual cross-over between natural and artificial systems. Von Neumann
and Turing were but two of the pioneers who contributed to the emergence of
bio-computing -- the extraction of computational principles or methods from
natural, biological systems.

Biologically-inspired computational methods such as artificial neural networks
and genetic algorithms have been successfully used to solve a wide range of
problems in the engineering domain. We can think of such methods as comprising
the ``first generation" of biocomputing techniques, in that they rely on
(often very) crude approximations of the fundamental underlying biological
principles (for example, the basic crossover operator used in genetic
algorithms). 

Such crude models have, up until now, been accepted for several reasons: the 
first is that they have produced solutions that are considered ``good enough"
in terms of their fitness for purpose. The second reason is borne out of
necessity, in that a sufficiently-detailed description or understanding of
the underlying biological system has, so far, eluded us. Both reasons for
acceptance of the {\it status quo} are now, we believe, beginning to be eroded,
both by a growing unease at the limitations of current nature-inspired models,
and by the speed at which our understanding of biology is increasing. We believe
that the time is right for the development of a ``second generation" of
bio-computational methods that draw much more closely on the growing
understanding of their biological inspiration. 

One possible explanation for the lack of recent progress 
in nature-inspired computing may be that the respective disciplines parted
company far too prematurely, As the report of a 2001 EPSRC workshop minuted,
``One of the major challenges facing collaborative work in biologically
inspired computing systems is the temptation to diverge far too early. This
has been the case to some extent in genetic algorithms where this method has
not moved forward as hoped because of premature divergence of the computing
and biology communities"\cite{bics}.

If the second generation of biocomputing is to emerge, it is perhaps the case
that a new convergence of disciplines is required, to re-ignite the initial
spark of interest that passed between them.
Developments in systems (and now {\it synthetic}) biology are driving advances
in biology, engineering and computer science -- crucially, these breakthroughs
are no longer unidirectional, in that expertise and concepts flow in a one-way
stream from one discipline to another. The new {\it systems-level} 
philosophy that is beginning to dominate 21st Century science dictates that
artificial boundaries between disciplines must be transcended or even
demolished. A first step in this process might be to revisit our perspective
on nature-inspired computing, and perhaps even reinvent it from the bottom-up.

% MA - changed its' to its

The UK research community has proposed a number of Grand Challenges for Computer Science research and ambitious plans for the development of a variety of research areas. Grand Challenge 7 (GC-7) \cite{1713} addresses the area of Non-Classical Computation, which includes exploring areas of both biologically inspired paradigms, and the exploitation of the natural world (for example, DNA computing and quantum computing) in order to develop new areas of computation. Part of its ambition is to bring together, once again, the disciplines that have prematurely drifted apart.   It has become clear that not all bio-inspired approaches are the same, each have their own contribution to make to the scientific community.  

% MA - changed "you" to "we" and slightly restructured rest of paragraph

In this position paper, we highlight a number of themes, the first of which is that {\em it is important to consider the type of biological system we wish to
exploit}, as different biological systems have different mechanisms that give
rise to different properties. The second, and the main thread in this position
paper, is that {\em high level abstractions of the underlying biology are no
longer sufficient}, and a deeper level of understanding is required.
It will become necessary to first {\it understand} and then {\it exploit}
the underlying richness of the biological system through an interdisciplinary
approach that combines biology with computation and engineering in a
synergistic manner. 

In order to expand on these themes, this paper will take the reader through two exemplar bio-inspired systems (section \ref{section:GPandAIS}), specifically genetic programming (GP) -- section \ref{subsection:GP} and artificial immune systems (AIS) -- \ref{subsection:AIS}. We will reflect on each of these themes as we describe how the paradigms have evolved over the years, and how they have begun (or otherwise) to address issues that these themes raise.  In this paper, we will also emphasise the importance to consider the deployment of artificial systems, i.e. is it a software or hardware system that is being constructed and deployed, as different considerations are required for each. We tackle this in section \ref{section:hardware} and again, reflecting on the themes outlined above in the context of evolvable hardware (EH).  Finally, in order to bring things together, we examine the area of natural genetic engineering (section \ref{section:cells} and \ref{section:NGE}) where we present a small case study of the development of a biologically inspired approach that is truly rooted in the biology of the system that we are trying to exploit, and, for a change, real feedback can be given to the biologists of insights into the biological system, rather than the wholesale pillaging that GP, AIS and EH have done to date.

\section{Capturing Evolution and Immunity} \label{section:GPandAIS}

\subsection{Genetic Programming}\label{subsection:GP}

Genetic Programming is a technique for breeding computer programs, loosely following the Darwinian theory of evolution \cite{koza92}. To be more specific, it is based on the new synthesis approach towards evolution, including genetic traits being inherited from generation to generation by way of molecular mechanisms based on DNA and its transformation into phenotypes \cite{bnkf98}. This approach followed in the footsteps of other quite successful adaptations of the evolutionary paradigm toward the solution of optimization problems with Genetic Algorithms \cite{holland75}, Evolutionary Programming \cite{fogel66} and Evolutionary Strategies \cite{rechenberg75}. In those earlier applications of the same fundamental idea, the substrate was somewhat different (fixed-length, fixed-representation data structures) and the goal was of a different nature (primarily a single optimum, kept constant over the entire run). Genetic Programming is different in that both the complexity of the task (be it behaviour of a robot, or a program or an algorithm) and the complexity of the data structure (for example, expressed in terms of the length of code) are variable over the course of the evolutionary run. It turns out that indeed both are tightly intertwined: There is no hope to find a solution to a task whose complexity might either be changing or might be unknown, without being able to adapt the complexity of the solution.

Early applications of Genetic Programming typically employed the so-called parse-tree representation, one of the clearest examples of an evolvable data structure. Later, a number of other data structures have been introduced which are able to evolve as well if not better than parse trees under circumstances. Prominent among them is the linear sequence of instructions of an imperative programming language \cite{banzhaf93,nordin94}, and the generic graph structure whose special case is the tree earlier used \cite{teller94}. After an initial flurry of publications on methods, during which different operators and selection methods were examined, besides data structures, the field has turned to application and exploitation of the method. Some spectacular applications published early paved the way for bold inquiries \cite{gritzhahn,sims,gruau,koza} and recently the field was emboldened to such a degree that applications rivalling and even superseding the performance of humans have been demonstrated. For instance, Koza has now applied for patent protection for one particular design generated by his highly parallel computational approach toward circuit design. Others have succeeded in designing antennae that will be sent into Space because they beat human designers on the same set of goals \cite{lohn:2004:GPTP}. An entire competition has sprung up, to be held regularly at the annual GECCO conference each year \cite{gecco} which is devoted to the products of GP (and other evolutionary methods) in producing designs being human-competitive or beating human performance. 

It might be asked in all fairness, whether this field has now come to maturity and merely needs to find its niche in the growing set of bio-inspired computational paradigms. We believe it has not, at least not yet. The reason is that during the same time as computer scientists and engineers were busy developing and exploring their methods for artificial evolution, biologist and other natural scientists made great strides at deciphering Nature's secrets and the success recipes of real evolution. In effect, the people taking inspiration were actually taking inspiration from a snapshot of a real system. Because of the explosion of activity in molecular biology, the explosion of knowledge about evolution and the fluidity of concepts, it was like shooting after a moving target, with all the difficulties accompanying the conquer of the unknown, with its twists and turns.  

We can expect a continued stream of new discoveries in Biology, that might translate into improvements of artificial evolutionary approaches in the future, provided we give them attention. As such, this field has not reached a steady state yet, but it is driven by further progress in one of the most dynamic human quests of this time. Furthermore, the enabling technology behind artificial evolution is large storage capacity and enormous computational power of the digital computer. These quantities are also quantities in transition, as attested by the continuous adherence to growth in both quantities according to Moore's law.

\subsubsection{Reflections on Genetic Programming}\label{subsubsection:GP} 
%JT ADDED 
{\bf Consequences of Exploiting Evolution} \newline \noindent

One of the key aspects of Genetic Programming is the adaptability of the genetic representation. This is limited, though, mainly to changes in the complexity of a proposed solution, as might be measured by the number of elements used (e.g. nodes in a graph or tree, lines of code in a linear representation). Early on, a curious phenomenon was discovered: The length of programs would tend to grow, and grow nearly without bounds, unless artificially restricted \cite{angeline94}. What was more, the resulting code would not always reflect more complexity in its behaviour. Upon closer inspection it was found, that so-called neutral code (early on called introns) \cite{Altenberg:1994EPIGP} was the main source of complexity growth of the data structures representing algorithmic behaviour \cite{Langdon}.

It was then found that this phenomenon, which the community agreed to call an emergent phenomenon of GP \cite{bnkf98} (a) was ubiquitous in runs, (b) had both positive and negative effects on resulting behavior, (c) was consuming many resources, and (d) needed to find an explanation from within the Genetic Programming method used. A substantial number of publications have in the meantime addressed this problem, and a consensus is beginning to emerge as to what the reasons are for the phenomenon. At the same time, remedies have been proposed to prohibit the effect (e.g., homologous crossover \cite{GPEM}), and other methods have been considered to introduce the same effect artificially (e.g., explicitly defined introns \cite{francone96}) . 

Following up in the tradition of other evolutionary computation approaches, the question of building blocks and the dynamics of their growth has been examined. For that to work, a number of preconditions had to be fulfilled, among them a better understanding of how the length-varying evolutionary algorithms work. An entire body of work has now been published on this question \cite{poli,stephens}, and our understanding how these algorithms actually work has been forwarded substantially.

Code-growth is not the only emergent phenomenon in Genetic Programming, it is perhaps the most obvious. Upon closer inspection of solutions bred by GP approaches, it was found that evolved code shows internal patterning reminding of the repetitive structure of natural genomes. Examined first in linear sequences of code \cite{complexsystems05}, it has been now confirmed in other representations as well \cite{eurogp05}. Much as in its natural counterpart, it seems to be again a phenomenon that has the aspects (a) to (d) mentioned above.

Exciting new developments have been reported by merging approaches from GP with approaches from the Artificial Life community \cite{pushGP} in recent years. The application sphere of GP (and other evolutionary approaches) is growing every year, and the amount of work published is growing correspondingly. Yet there is also an ñundergroundî stream of work that is not published or patented and instead held as trade secret. This refers to applications  mostly in the financial business where having a slight edge over the competitor might produce a big difference money-wise.

The number of applications for GP is in the hundreds, a quick look at publications from 2005 alone reveiles that those range from elementary particle physics \cite{NuclInstr} to bankruptcy prediction \cite{Lensberg}.

However, there are a number of areas where we have to expect further progress before GP can really make a difference in the world. First and foremost, present GP techniques do not scale well. That is to say, at present, useful applications are restricted to short programs of a complexity comparable to 50 to 250 lines of code. Otherwise, modularity needs to be engineered into the solutions, somewhat to the distraction of an unbiased evolutionary search process.  \newline 

\noindent{\bf Levels of Abstraction Employed - Getting Back to the Biology} \newline \noindent

In most of the human-competitive programs evolved so far, a structuring of the evolutionary process has turned out to be necessary this way or another. As Koza et al. have explained in their recent book \cite{kozaV}, automatically defined functions (ADFs) could be employed to allow such a structuring. Originally completely engineered, they now have more power to evolve under selection pressures.  

It has been argued that without inclusion of a developmental process that might enable the evolution of genome structuring and the upscaling of solutions through growth GP will continue to suffer from scalability and the lack of potential for modularity \cite{miller_banzhaf,Bentley,Hornby}. The inclusion of such a process, however, does not come free. It will need to considerably complicate the genotype-to-phenotype map if it should have prospects of improving the paradigm, something that practinioners are weary about. We believe, however, that without implementation of such a developmental structuring process which includes to consider regulation as an inherent part of evolution \cite{arn}, further progress will be stymied.

As has been discovered in molecular biology, there is a lot more going on with the genome than meets the eye. Although only 1-3\% of the human genome are translated into protein and thus consists of genes, more than 50\% of the genome is transcribed \cite{kapranov}. This indicates that we have barely started to understand the formation of phenotypes from genetic information \cite{conserved}. Is transcription a more important step in the formation of the phenotype than translation into proteins?

In addition, phenotypic differences can come about without a difference in the genetic makeup of an individual \cite{wong}. This points to the dimension of Epigenetics as of primary importance if we want to understand evolution \cite{meaney}. How could those effects be incorporated into Genetic Programming,
or should we say Epigenetic Programming?   

As we can see, even in the world of evolution things are not as they used to be. Thus, we would do well in continuing to explore the diversity of Nature, and to apply various methods in various combinations. Just as computer hardware is becoming more ubiquitous and seemlessly connected to biological systems because technical progress has allowed integration, so should our algorithms become 
more life-like and compatible with human behavior, because we look at a closer
modelling of biological behaviour. Evolution, however, is not the only adaptive
system we'd have to study. The neural system, the endocrine system and the immune system need to be studied, too. The next section discusses one of them
that has recently become the focus of attention of researchers in bio-inspired
computing.

\subsection{Artificial Immune Systems}\label{subsection:AIS}
The immune system is a very complex system that undertakes a myriad of tasks.  The abilities of the immune system have helped to inspire computer scientists to build systems that mimic in some way, various properties of the immune system \cite{deCastroLN_2002a}.  This field of research, Artificial Immune Systems (AIS) has seen the application of immune inspired algorithms to problems such as robotic control \cite{Krohling:2002:ICARIS}, network intrusion detection \cite{Forrest1997a,kim-thesis:2002}, fault tolerance \cite{Canham:2002:ICARIS,ayara2005}, bioinformatics \cite{nicosia-thesis:2004} and machine learning \cite{maria,WatkinsA_2004a}, to name a few. From a computational point of view, the immune system has many desirable properties that could be endowed on artificial systems.  These properties include: robustness, adaptability, diversity, scalability, multiple interactions on a variety of timescales and so on.  

\subsubsection{ A Brief History of Artificial Immune Systems}
The origins of AIS has its roots in the early theoretical immunology work of Farmer, Perelson and Varela\cite{FarmerJD_1986a,perelson:1989,Varela:1988} in which a number of theoretical immune network models were proposed to describe the maintenance of immune memory in the absence of antigens.  These models, whilst controversial from an immunological perspective, began to give rise to an interest from the computing community.  The most influential people at crossing the divide between computing and immunology in the early days were Hugues Bersini and Stephanie Forrest.  In the case of Hugues, he attended a talk by Francis Varela in 1985, and Hugues made the decision there and then to begin working with Varela. In the case of Stephanie Forrest, she happened to be car sharing on the way to work with Alan Perelson, and thus their working relationship began there.  It is fair to say that some of the early work by Bersini \cite{BersiniH_1991a,Bersini-92} was very well rooted in immunology, and this is also true of the early work by Forrest \cite{ForrestS_1994a,Hightower-95}. All of these works formed the basis for a great deal of excellent foundational work in the area of AIS. In the case of Bersini, he concentrated on the immune network theory, examining how the immune system maintained its memory and how one might build models and algorithms mimicking that property.  With regards to Forrest, her work was focussed on computer security (in particular network intrusion detection) \cite{Forrest1997a,Hofmeyr:2000} and formed the basis of a great deal of further research by the community on the application of immune inspired techniques to computer security.

At about the same time as Forrest was undertaking her work, researchers in the UK started to investigate the nature of learning in the immune system and how that might by used to create machine learning algorithms \cite{cook:1995}.  Initial results were very encouraging, and they built on their success by applying the immune ideas to the classification of DNA sequences as either promoter or non-promoter classes,  \cite{hunt:1996} and the detection of potentially fraudulent mortgage applications \cite{hunt:1998}. 

 This then spawned more work in the area of immune network based machine learning over the next few years, notably in \cite{jonthesis,1121} where the Hunt and Cook system was totally rewritten, simplified and applied to unsupervised learning (very similar to cluster analysis).  This thread of work on machine learning, spawned yet more work in the unsupervised domain, but trying to perform dynamic clustering ( where the patterns in the input date move over time).  This was met with some success in works such as \cite{Wierzchon:2002:ICARIS,Neal:2002:ICARIS}. At the same time, using other ideas than the immune network theory, work by \cite{Hart:2002:ICARIS} used immune inspired associative memory ideas to track moving targets in databases.  
 
 In the supervised learning domain, very little happened until work by \cite{andrewthesis} developed an immune based classifier known as AIRS.  The system developed by Watkins (and later augmented in \cite{WatkinsA_2004a}) and then turned into a parallel and distributed learning system in \cite{watkins:phd}, has shown itself to be one of the real success stories of immune inspired learning \cite{goodman2,goodman,watkinsclonal}. 

 Of course, there was other activity in AIS at this time, machine learning was one small area.  To outline all the applications of AIS and developments over the past 5 years would take a long time, and there are some good review papers in the literature, and the reader is directed towards those \cite{DasguptaD_1999a,timmis:dm,deCastroLN_2002a,GarrettSM_2005a,HartE_2005a}.
 
 The International Conference on Artificial Immune Systems (ICARIS) conference series was born in 2002 and has operated in subsequent years \cite{TimmisJ_2002a,TimmisJ_2003a,NicosiaG_2004a,icaris05}. This is the best reference material to read in order to grasp the variety of application areas of AIS, and also the developments in algorithms and the more theoretical side of AIS. 

\subsubsection{Reflections on Artificial Immune Systems} \label{subsubsection:reflectAIS}

\noindent{\bf Consequences of Exploiting Immunology}  \newline \noindent

Taking the early work of Forrest et al \cite{forrest1994} as an example, their work led to a
great deal of research and proposal of immune inspired anomaly
detection systems \cite{Forrest1997a}.  Results reported in these
works, did hint at the possibility that the immune approach was going to be beneficial
to some degree, as work showed that both known and novel intrusions could be detected.
However, given the typical representation used (binary), the  {\em r-contiguous bits} matching
rule was typically used to compare contiguous regions of the binary string, there were issues with computational efficiency. The {\em r-chunk}
rule, developed later, made it computationally more efficient to generate a set of
detectors of the non-self space (in hamming shape space) and later
computationally more efficient methods were developed in real-valued
shape space \cite{GonzalezFA_2003a}

Unlike GP (section \ref{subsubsection:GP}) where code bloat was a problem, in this case the problem was be able to generate enough detectors capable of covering the space effeciviy;  In essence, what ended up occuring was an exponential relationship between the size of the {\em self} data and the number of detectors that could be  generated.  Work in \cite{StiborT_2004a,StiborT_2005a} presented an in-depth theoretical
analysis of the negative selection algorithm over real and hamming
shape spaces. The authors suggest that over the hamming
shape-space, the approach is not well suited for real-world anomaly
detection problems. Problems seem to arise with the generated detector set, under-fits the training data, exponentially for small values of $r$ (where $r$ is the size of the chunk. They suggest that in order avoid this
under-fitting behavior, the matching threshold value $r$ must lie near
$l$ (the length of the string). However, they also point out that this has
a consequence.  This is that the detector generation process is once
again infeasible, since all proposed detector generating algorithms
have a runtime complexity which is exponential in $r$. In addition to
their theoretical arguments, they undertook a simple study of
comparison between the negative selection approaches on a one-class
support vector machine (SVM).  When comparing the
work of \cite{GonzalezFA_2003a}, (the real-valued negative selection algorithm
with variable-sized detectors) results revealed, that the
classification performance of the method not only crucially depended
on the size of the variable region, but results from the one-class SVM
provides as good, if not better results.   As argued by \cite{HartE_2005a}, it is not clear that AIS to date, has made a real breakthrough in the ''natural'' application of itself to network security.

Another example, is the work of Timmis et al \cite{timmisandneal2001}.  In \cite{timmisandneal2001}, an immune network inspired algorithm was proposed that was capable of performing unsupervised learning. Initial results were very encouraging, but further investigations in \cite{KnightT_2001a} highlighted a number of problems with that work, and identified a different behavioral pattern to the original work. The subsequent investigation discovered that the algorithm would naturally discover the strongest pattern within the data set that it was applied to, and the network would effectually converge into a single cluster: not the desired behavior at all. The pressure within the network was too great for weaker cells to survive, which came about by a naive implementation of the network interactions where stimulation and suppression between cells was not balanced effectively.  As pointed out by \cite{garret:arpen}, a too simplistic approach had been taken in the representation of the data vectors and in particular with the definition of their interactions.  \newline

\noindent{\bf Levels of Abstraction Employed - Getting Back to the Biology} \newline \noindent

The original AIS were with an interdisciplinary slant, taking care to develop algorithms that were ''faithful'' to their immunological roots. For example, Bersini \cite{BersiniH_1991a,Bersini-92,Bersini:1994} pays clear attention to the development of immune network models, and then applies these models to a control problem characterised by a discrete state vector in a state space. BersiniÕ's proposal relaxes the conventional control strategies, which attempt to drive the process under control to a specific zone of the state space; he instead argues that the metadynamics of the immune network is akin to a meta-control whose aim is to keep the concentration of the antibodies in a certain range of viability so as to continuously preserve the identity of the system.  

There are other examples of interdisciplinary work, such as the development of immune gene libraries and ultimately a bone marrow algorithm employed in AIS \cite{Hightower-95}, and the development of the negative selection algorithm and the first application to computer security \cite{forrest1994}. However, in more recent years, work on AIS has drifted away from the more biologically-appealing models and attention to biological detail, with a focus on more engineering-oriented approach. This has led to systems that are examples of ''reasoning by metaphor'' \cite{stepney2004}. These include simple models of clonal selection, immune networks and negative selection algorithms as outlined above. For example, the clonal selection algorithm (CLONALG) \cite{clonalg}, whilst intuitively appealing, lacks any notion of interaction of B-cells with T-cells, MHC or cytokines.  In addition, the large number of parameters associated with the algorithm, whilst well understood, make the algorithm less appealing from a computational perspective.  aiNET, again, whilst somewhat affective, does not employ the immune network theory to a great extent.  Only suppression between B-cells is employed, whereas in the immune network theory, there is suppression and stimulation between cells. With regards to negative selection, the simple random search strategy employed, combined with using a binary representation, makes the algorithm computational so expensive, that it is almost unusable in a real world setting \cite{Stibor:2005b}.

However, in the past year or so, work by the Danger Team \cite{dangerteam} has started to address this imbalance. For example, recent work by \cite{greensmith:2005} has begun initial explorations into the use of dendritic cells (which are a type of cell found in the innate immune system, as a mechanism for identifying dangerous (or anomalous) event in a data stream.  Whilst that work is still preliminary and works only on static data at the moment, there is a great deal promise there, and may go some ways towards making a real breakthrough in the intrusion detection area of AIS research.  Work linked to that is by \cite{bentley:2005} proposes an artificial tissue, which is a type of representation of the data space that can evolve and adapt over time. Again, this is very preliminary work, but could prove useful bridge between the data and the immune algorithm itself.  In addition to this, work in \cite{concept2005} proposes a ''conceptual framework'' for the development of AIS (although, it could be generalised to any bioinspired approach). They propose a greater interaction  between computer scientists, engineers, biologists and mathematicians, to gain better insights into the workings of the immune system, and the applicability (or otherwise) of the AIS paradigm will be gained. These interactions should be rooted in a sound methodology in order to fully exploit the synergy. 

Whilst the immune system  is clearly an interesting system to investigate, if viewed in isolation, many key emergent properties arising from interactions with other systems will be missed.  Such systems do not operate in isolation in biology, therefore, consideration should to be given to the interactions of the immune, neural and endocrine systems, and how, together, they allow for emergent properties to arise \cite{endo1,endo2,Sieburg:1991:CDM}.  Immune, neural and endocrine cells express receptors for each other. This allows interaction and communication between cells and molecules in each direction. It appears that products from immune and neural systems can exist in lymphoid, endocrine and neural tissue at the same time. This indicates that there is a bi-directional link between the nervous system and immune system. Therefore, it would seem that both endocrine and neural systems can affect the immune system. There is evidence to suggest that by stimulating areas of the brain it is possible to affect certain immune responses, and also that stress (which is regulated by the endocrine system) can suppress immune responses: this is also reciprocal in that immune cells can affect endocrine and neural systems. The action of various endocrine products on the neural system is accepted to be an important stimulus of a wide variety of behaviours. These range from behaviours such as flight and sexual activity to sleeping and eating \cite{breach}.

Computationally then, what does this have to say to AIS?  It should be possible to explore the role of interaction between these three systems. One interesting avenue would be to design an AIS to help select the types of components which will be most useful when added to a control system at any moment (differentiation) and to remove components when they are proving harmful to the control system (apoptosis). The biological immune system cells select which action to perform by detecting properties of the cells and chemical environment through molecular interactions at membrane receptors. In an artificial system, similar properties can be detected by looking at activation states of artificial neurons and endocrine cells as well as global state information such as current consumption and battery levels. Thus the artificial immune system components can make similar decisions within the artificial context.

\section{What about the Medium?} \label{section:hardware}

%jt added
We have now reviewed two bioinspired paradigms, one well established (GP - section \ref{subsection:GP}) and one relative newcomer, AIS - section \ref{subsection:AIS}).  As you may, or may not, have noticed, there has been little mention as to the nature of the system on which they are developed.  In this section, we review the area of evolvable hardware (EHW), and how evolution has been used in hardware systems and the considerations that should be taken into account when developing such systems.

Evolvable HardWare (EHW) is a method for electronic circuit design that uses inspiration from biology to evolve rather than design hardware. At first sight this would therefore seem very appealing. Consider two different systems, one from Nature and one human engineered: the human immune system (Nature design) and the computer that we are writing this article on (human design). Which is more complex? Which is most reliable? Which is optimized most? Which is most adaptable to change? The answer to almost all of those questions is most likely the immune system. One could argue that the only winner from the engineered side could be the question ''Which is optimized most?'' From this, the obvious conclusion might be to only use methods based on the way Nature works, and not use the myriad of current ''human'' design methods. However, it is not quite that straight forward. Nature has a number of advantages over current engineered systems, not least of which are time (most biological systems have been around for quite a long time) and resources (most biological systems can produce new resources as they need them, eg new cells).

However, this does not mean that evolutionary designs canÕt be useful. As a simple example where evolutionary techniques can have a real benefit, consider the growing issues associated with keep electronic systems running and operational for long periods of time, often in unpredictable environments.

Since the basis of evolutionary algorithms is a population of different systems which compete for the chance to reproduce, it already contains a type of redundancy, since each system in the population is different. When a fault occurs in the system, either through external conditions (sensors failing or giving incorrect readings), or through internal electronic faults, it will affect different systems in different ways due to the diversity of solutions. This should mean that particular individuals will not be effected by the fault and therefore provide tolerance to the fault.

Evolvable hardware is a new field that brings together reconfigurable hardware, artificial intelligence, fault tolerance and autonomous systems. Evolvable hardware refers to hardware that can change its architecture and behaviour dynamically and autonomously through interaction with its environment. Ideally this process of interactions and changes should be a continuous one and should be open-ended.

Evolved hardware is a new field that brings together reconfigurable hardware, artificial intelligence, fault tolerance and autonomous systems. Evolved hardware refers to hardware that has been created through a process of continued refinement and, where the evolutionary process will terminate when a sufficiently ''good'' individual has been found.

Both evolvable and evolved hardware make use of evolutionary techniques in order to produce a system that performs to some specification; both are to some extent autonomous, and both may have properties that endow the final system limited fault tolerance. The major difference, but not the only one, between the two is that evolved systems do not change after a good individual has been found, and are therefore rather static designs (as are most human designs). Evolvable systems continue to have the possibility of changing throughout their existence. Hence, as systems change (eg components fail), as the environment changes (temperature changes, sensor failures etc) an evolvable system will be able to adapt to these changes and continue to operate effectively.

\subsection{A Brief History of Evolutionary Hardware Systems}
It could be argued that artificial intelligent system design should address features such as autonomy, adaptability, robustness, and fault-tolerance. Autonomous robot navigation in dynamic environments represents a very challenging task, which needs to take into account such features. Conventional approaches based on off-line learned control policies generally do not work appropriately when implemented in real time environments. For example, the actual hardware system implementing the evolved behaviour may well not accurately match the simulation environment used during evolution. The sensors and actuators used in the real system may have different characteristics to those used in the simulation (e.g. IR sensors in a bright environment would operate differently to those in a dark or changing light environment). The development of a new research field named Evolvable Hardware (EHW), that is application of evolutionary algorithms \cite{fogel:1995} to automatic design and or reconfiguration of electronics circuits \cite{Zebelum:2001}, presents a promising approach to tackle the problem of adaptation in unknown/changing environments.

Two methodologies have been established for the design of EHW: Extrinsic and intrinsic \cite{thompson:1996,miller:1998,layzel:1998,haddow:2000,hollingworth:2000}. In the former case, both the evolutionary process  as well as the fitness evaluation of each individual (the circuit) is simulated in software. The entire design is undertaken off-line and once the evolutionary process has completed, the ''best'' member of the final population is downloaded onto the hardware. In the latter case, the evolutionary process maybe executed in software but each individual is executed and evaluated in hardware \cite{haddow:2000,hollingworth:2000}. Developments in electronic devices such as Field Programmable Gate Array (FPGA), which are reconfigurable devices with no pre-determined function \cite{shir:1996}, have enabled theoretical ideas of intrinsic evolution to become a reality in the last few years. Each individual is represented as a bit string (genotype) that is downloaded to the chip as configuration data. This data includes a definition of each cellÕs functionality as well as the topology of the system.

HiguchiÕs group in Japan have taken a different approach to EHW. Rather than use extrinsic evolution, or intrinsic evolution on COTS, they have developed a single LSI chip that is sepcifically designed to support evolvability \cite{iwata:2001}. They have developed a gate-level chip that consists of genetic algorithm hardware, reconfigurable hardware logic and the control logic required. This chip, and variants of it, have been applied to a number of applications, including: artificial hand controller, autonomous robotics, data compression of image data, analogue chip design for cellular phones, optical system adjustment and adjustment of clock timings \cite{higuchi:2003,keymeulen:1999,liu:2000}. What these results show is an increased cycle time in the evolutionary process, due to the specialised optimisation that has taken place in the hardware.

Evolutionary computation (e.g. Genetic Algorithms (GA), Evolutionary Strategies (ES) and Genetic Programming (GP)) have been applied to EHW, for example using a binary representation, appears to be convenient since it matches perfectly with the configuration bits used in FPGAs. There are however, huge problems associated with the evolution of large circuits (or actually what today are probably considered small circuit designs) due to problems of scaling. That is, with direct genotype-phenotype mappings such as these as the circuit complexity increases so does the size of the genotype and the size of the search space. A number of papers have been published to evolve on-line FPGA-based robot controllers using these methods \cite{thompson:1995,keymeulem:1997,haddow:1999,tan:2002} using COTS. One of the main problems of evolving on a FPGA is the Genotype-Phenotype mapping. Effective methods to solve this problem using intrinsic EHW is proposed.  In intrinsic EHW, the fitness is evaluated on target hardware. Therefore changes in environment are reflected immediately in the fitness evaluation. For example, the problem of adaptation of autonomous robot navigation in changing environments consists of finding a suitable function F (the controller) which maps the inputs from sensors to the outputs (control signal to the motors).

GP is becoming more popular in EHW. Techniques such as Cartesian Genetic Programming (CGP) \cite{miller:2000} and Enzyme Genetic Programming (EGP) \cite{lones:2002} have been applied extensively to EHW. A recent development, and an attempt to move back to biology is considered in \cite{cai:2005}.

A criticism of
CGP (and GP in general) is that the location of genes within the chromosome has a
direct or indirect influence on the resulting phenotype. In other words, the order in
which specific information regarding the definition of the GP is stored has a direct or
indirect effect on the operation, performance and characteristics of the resulting program.
Such effects are considered undesirable as they may mask or modify the role of
the specific genes in the generation of the phenotype (or resulting program). Consequently,
GPs are often referred to as possessing a direct or indirect context representation.

An alternative representation for GPs in which genes do not express positional dependence
has been proposed by Lones and Tyrrell \cite{lones:2002}. Termed implicit context representation,
the order in which genes are used to describe the phenotype (or resulting
program) is determined after their self-organised binding, based on their own characteristics
and not their specific location within the genotype - much more like biology. The result is an implicit
context representation version of traditional parse-tree based GP termed Enzyme Genetic
Programming. The authors have since implemented an implicit context representation
of CGP, termed Implicit Context Representation Cartesian Genetic Programming
(IRCGP), specifically for the evolution of image processing filters \cite{cai:2005}.

In many evolutionary algorithms, whether intrinsic or extrinsic, once the final criteria has been met (that might be the required fitness level, or the maximum number of generations has been reached) the evolutionary process stops, and the best of population is used in the implementation. An alternative approach is to allow continuous evolution throughout the lifetime of a system. The evolutionary process continues, however, once a member has been chosen for implementation the evolutionary process does not stop. Such a continuous process allows a system to be more responsive to environmental changes. For example, evolution can cope with errors during runtime. The fitness might drop at the instant the error is activated, but the evolutionary process autonomously deals with this drop in fitness and recovers back to an acceptable fitness level, and hence acceptable level of functionality over a number of generations.

\subsection{Reflections on EHW}

\noindent{\bf Consequences of Exploiting Evolution} \newline \noindent

Original design of digital circuitry is an area where the EHW hardware community (which tended to be, and in many case still are, the main focus of EHW) has had limited success breaking into. To fully appreciate why, it is important to understand how industry does digital design.

The overwhelming majority of digital design today is done using electronic design automation (EDA) tools. Complicated designs are usually implemented in an FPGA or in an application specific integrated circuits (ASIC) where the device density permits high-speeds in small packages. The complexity of these designs makes hand design methods impossible. EDA tools automate the design process thereby reducing design time, improving yields, and reducing nonrecurring costs.

The process begins by describing the digital circuit in a computer program written in a hardware description language (HDL). The two most ubiquitous HDLs are Verilog and VHDL, both of which are specified by IEEE standards. This design can be expressed in a mixture of different levels of abstraction. The compiled source code is input to a synthesiser along with a component library and any design constraints on timing or power consumption. The synthesiser is responsible for taking the design described by the HDL and, using devices from the component library, creating a circuit that satisfies any design constraints.

The output of the synthesizer then goes to another EDA tool that assigns logic functions to configurable logic blocks inside an FPGA and determines the routing between blocks. A bitstream produced by the design implementation tool is used to physically program the FPGA. Verification of the design occurs at various places in this design process. The HDL description is simulated to verify the design meets all functional specifications.

A second simulation is performed after synthesis to ensure the synthesised circuit still functions properly. Once the synthesised circuit design is placed and routed, thorough timing analysis is conducted. Finally, the programmed FPGA is placed into its target system for a full functional and timing check. If the verification fails at any stage, the original HDL description must be modified and the synthesis process repeated. For example, if the timing analysis identifies a flaw, the designer could describe the design at a lower level of abstraction, which allows for a tighter control over what gets synthesised.

EHW practitioners need to understand the FPGA design flow is in place and widely used throughout the integrated circuit industry today. In essence, this means the EHW community has to show substantial, significant advantages over an established method before making any real inroadsÑand that presents several challenges to any EHW method trying to perform circuit synthesis. 

An additional, but fundamental issue all EHW users have to face is scalability. A typical ''big'' EHW system might be a few 100 transistors (1970Õs technology for the microprocessor manufactures), current chip designs are more like 100,000,000 transistors. \newline

\noindent{\bf Levels of Abstraction Employed - Getting Back to the Biology} \newline \noindent

The area of adaptive system design and control is potentially where EHW methods have the greatest potential for digital, analog and mixed signal systems. Circuitry can be adapted, i.e., reconfigured, to take on new roles or for fault recovery. Consequently, the focus should be on adaptation for fault recovery operations.

Circuits are adapted in-situ. What makes such an environment particularly difficult to work in is the user almost never has complete knowledge about why the original circuitry failed. Obviously faults can degrade a circuitÕs performance, but any change in the operational environment can do this as well.

Regardless of the cause of a fault, reconfiguration done in-situ is especially challenging for two reasons: (1) faults can be hard to detect and isolate, and (2) the reconfiguration function itself may be compromised by the fault. However biological systems seem to operate for the majority of their time with all of these issue and more continually challenging them.

At the present time a significant portion of EHW-based fault recovery investigations rely on simulations and usually fairly simplistic models of biological evolution. As with many evolutionary-type systems, genotype to phenotype mappings tend to be rather simplified, usually one-to-one. As mentioned already developmental processes are seldom considered (although as already pointed out again, this can lead to its own issues). Investigations are needed to develop intrinsic evolutionary methods for autonomous systems with limited resources. This should also include recovery techniques. Like biological systems these evolutionary methods should not stop when one good solution is found. We need to consider how we can incorporate biological open-ended evolution into our systems?

More work needs to be done on developing EAs that can intrinsically evolve circuit configurations with imprecisely defined performance objectives (imprecise in the sense one does not know what level of performance can be achieved). Evolving benign configurations where further damage is contained and controlled is of interest. Again more accurate models of the equivalent biological processes need to be considered for our hardware systems.

Studies are needed to determine how effective EHW-based recovery methods are when the computing resources they run on are degraded by environmental conditions. Can we somehow use homeostasis-type ideas to make our systems adapt intelligently to ensure critical functions are kept operational, at the expense of other, less critical, functions - for example, the immune-endocrine-neural system (section \ref{subsubsection:reflectAIS}). Many fault recovery scenarios involve injecting arbitrary faults into an operating circuitÑe.g., a randomly chosen switch in an FPTA is forced open. It is not clear if such a fault is likely to occur in isolation or whether it results from some other fault. This ambiguity leads to the development of recovery methods that may have limited usefulness. A major issue with all fault-tolerant systems is error detection. This is mostly ignored in EHW systems. Artificial Immune Systems should have a role to play here and more should be made of the combination of multiple bio-inspired ideas: EHW + AIS, AIS + ANN + endocrine etc., as we discussed in section \ref{subsubsection:reflectAIS}

\section{At the Interface of Biology and Computing}\label{section:cells}

The previous sections have already shown in detail how living organisms may 
easily be thought of as information processing systems. Biological systems
such as individual cells are capable of performing amazingly intricate and
sophisticated information processing tasks, including pattern recognition and
distributed memory storage (previously described in the
context of the immune system), 
pattern {\it formation} \cite{bb91}, distributed communication
\cite{benjacob04} and adaptive control \cite{sontag04}.

As we have already seen, descriptions of 
cellular systems may be usefully abstracted and applied to the solution of
human-defined computational problems. In particular, studies of bacterial 
attraction and movement have been successfully applied to (among other problems)
the training of artificial neural networks \cite{delgado} and the design
of aircraft aerofoils \cite{muller}. In addition, actual living cells have also
been directly {\it engineered} to perform simple computational tasks. In 1999, Weiss
{\it et al}. \cite{weiss99} described a
technique for mapping digital logic circuits onto genetic regulatory
networks such that the resulting chemical activity within the cell
corresponded to the computations specified by the desired digital circuit.
There was a burst of activity in 2000, when two papers appeared in
the same issue of {\it Nature}, both being seminal contributions to
the field. In \cite{elolei00}, Elowitz and Leibler described the
construction of an oscillator network that caused a
culture of {\it E.coli} to periodically glow by expressing a fluorescent
protein.  Crucially, the period of oscillation was slower than the cell
division cycle, indicating that the state of the oscillator was
transmitted from generation to generation. In \cite{gacaco00},
Gardner {\it et al.} implemented a genetic toggle switch in {\it E.coli},
the switch being flipped from one stable state to another by either
chemical or heat induction.
These ``single cell" experiments demonstrated the feasibility of
implementing artificial logical operations using genetic
modification. In \cite{sav01}, Savageau addresses the issue of
finding general design principles among microbial genetic
circuits, citing several examples. Several more examples of
successful work on cellular computing may be found in
\cite{Amos04}.

It is clear, therefore, that working at the interface of cellular biology and 
engineering/computer science can generate tangible benefits in terms
of ``real world" applications. However, these applications do not bring us any
closer to a real break-through in terms of a completely novel computational
paradigm. Moreover, with very few exceptions, it is rarely the case that such
studies add anything to our overall understanding of the underlying biological
system. Here, we argue that the development of novel biological algorithms
should be, at least in part, motivated by a desire for longer-term insights, %jt added
and not just short-term applications.  Given the work presented so far in sections \ref{subsection:GP} and \ref{subsection:AIS}, there seems to be clear evidence that no matter how appealing creating applications can be, there are some inherent difficulties in adopting biologically inspired approaches, and it may be possible to try and circumvent some of these, through a more rigorous and in-depth investigations.

In the following section, we further strengthen
the main overall argument of this article -- that simple abstractions, whilst
useful, are limited, and that one needs to consider in detail the
underlying biology if such work is to have long-term general significance.
We support our argument by reviewing recent work on {\it natural genetic
engineering}. This relatively novel -- and still controversial -- view of
evolution, proposed mainly by Jim Shapiro, centres on the ability of
individual cells to restructure their own
genomic information in response to reproductive pressures. A deeper 
understanding of the fundamental underlying processes will benefit
not only biologists attempting to gain new insights into evolution, but
computer scientists and engineers seeking to use nature as the inspiration for
robust and adaptable hard/software systems. Crucially, though, this
investigation poses twin challenges to both biologists and computer scientists,
and neither community will succeed in isolation. However, the anticipated
benefits are wide-ranging and profound. As Shapiro himself argues: ``These
challenges should be high on the research agenda for the 21$^{st}$ Century. It
is likely that meeting them will lead us to new computing paradigms of great
creative power, like evolution itself" \cite{shapiro05}.

\section{Natural genetic engineering} \label{section:NGE}

Only relatively recently has the view of the genome changed from that of a
collection of relatively independent genetic units, to that of a {\it system},
made up of an organised collection of interacting and interdependent modules
\cite{shapiro99}. The coding regions of genes provide the templates for
proteins, which are generated when the gene is {\it expressed}, and these
proteins can themselves affect the expression of other genes \cite{jacmon61}.
Shapiro argues that the genome has a precise 
architecture encoded by the distribution of various non-coding repetitive DNA
sequences (often erroneously referred to as ``junk DNA"). As this system-level 
architecture governs the ``day to day" functioning of the cell, it follows
that alterations to this structure may well be much more significant in terms
of evolution than modifications to individual proteins \cite{shapiro99}.
By ``cutting and splicing" their own DNA, organisms may therefore
reorganise both their repetitive and their coding
sequences to generate new functional genomic systems.
{\it Natural genetic engineering} (NGE) \cite{shapiro97,shapiro05} is the term given
to this ability or capacity of organisms to modify or reorganise their own
genome in response to certain pressures. As we have seen in Section \ref{subsection:GP}, the shuffling of interdependent program modules is a very
effective computational strategy when combined with some sort of selection
pressure. This reorganisation may occur at different time-scales, and for a
multitude of reasons. For example, it is clear that NGE occurs, at an
intermediate time-scale, in the immune system \cite{shapiro05}.  Here, 
NGE progresses over the course of multicellular development, with cellular
differentiation providing the system ``clock". The notion of timescales, and the variety of them in the immune system, would seem to have been missed by the vast majority of AIS to date.

The problem
of encoding a virtually infinite array of response molecules given a finite
coding sequence region appears to have been solved by immune system lymphocytes
utilising NGE. Here, sequences of controlled DNA rearrangements generate
novel protein-coding regions that are used to generate new antigen-binding
molecules. Moreover, a ``real time" positive feedback loop amplifies the 
cells that have succeeded in generating molecules with a sufficient ``fit" 
to the antigen, and then these cells undergo a further process of DNA
``tweaking" to further increase specificity. This gives
lymphocytes an extraordinary degree of responsiveness; they have evolved to
{\it themselves} evolve rapid and specific adaptations
\cite{shapiro99,shapiro05}.

It is clear that NGE also occurs at a much more rapid time-scale than
that of lymphocyte differentiation. Many organisms are capable of extremely
rapid genomic rearrangement, perhaps the most striking example being that
of the {\it ciliates}.

\subsection{NGE in ciliates}

{\it Ciliate} is a term applied to any member of a group of around
10,000 different types of single-celled organism that are
characterized by two features: the possession of hair-like {\it
cilia} for movement, and the presence of two kinds of {\it nuclei}
instead of the usual one. One nucleus (the {\it micronucleus}) is
used for sexual exchange of DNA, and the other (the {\it
macronucleus}) is responsible for cell growth and proliferation.
Crucially, the DNA in the micronucleus contains an ``encoded"
description of the DNA in the working macronucleus, which is decoded
during development. This encoding ``scrambles" fragments of the
functional genes in the macronucleus by both the permutation (and
possible inversion) of partial {\it coding} sequences and the
inclusion of $non$-coding sequences.

It is the macronucleus (that is, the ``housekeeping" nucleus) that
provides the RNA ``blueprints" for the production of proteins. The
micronucleus, on the other hand, is a dormant nucleus which is
activated only during sexual reproduction, when at some point a
micronucleus is converted into a macronucleus in a process known as
{\it gene assembly}. During this process the micronuclear genome is
converted into the macronuclear genome. This conversion reorganizes
the genetic material in the micronucleus by removing noncoding
sequences and placing coding sequences in their correct order. This
``unscrambling" may be interpreted as a computational process.

The exact mechanism by which genes are unscrambled is not
yet fully understood. We first describe experimental observations
that have at least suggested possible mechanisms. We then describe
a computational model of the process. We conclude this Section with a discussion
of the computational and biological implications of this work.

\subsection{Biological background}

The macronucleus consists of millions of short DNA molecules that
result from the conversion of the micronuclear DNA molecules. With
few exceptions, each macronuclear molecule corresponds to an
individual gene, varying in size between 400 b.p. ({\it base pairs})
and 15,000 b.p. (the average size is 2000 b.p.). The fragments of
macronuclear DNA form a very small proportion of the micronucleus,
as up to 98$\%$ of micronuclear DNA is noncoding, including
intergenic ``spacers" (that is, only $\sim2\%$ of the micronucleus
is coding DNA), and all noncoding DNA is excised during gene
assembly.

\subsection{IESs and MDSs}

\begin{figure}
\begin{center}
\includegraphics[width=2.5in]{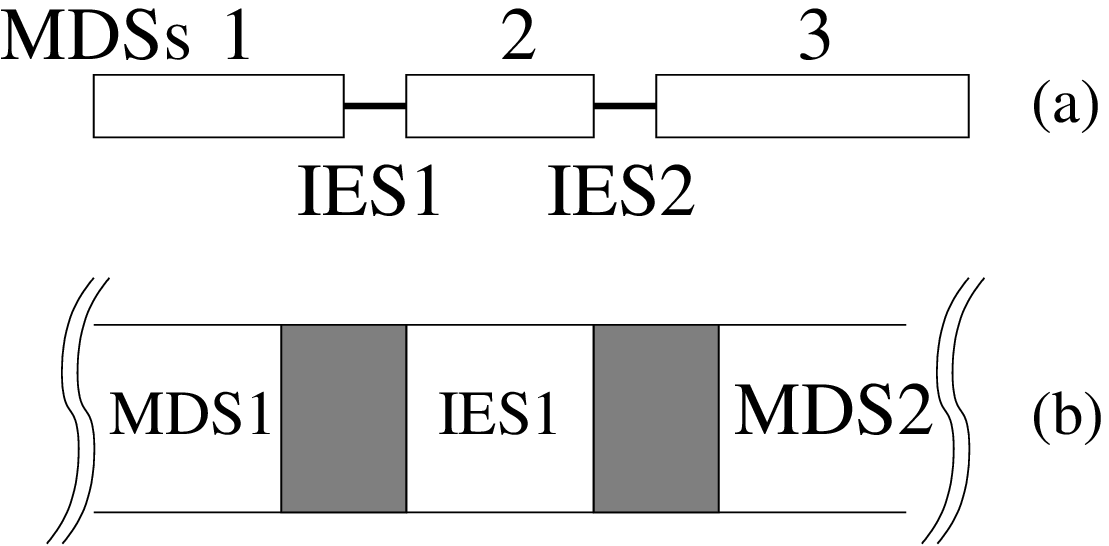}
\caption{(a) Schematic representation of interruption of MDSs by
IESs. (b) Repeat sequences in MDSs flanking an IES (the outgoing
repeat sequence of MDS1 is equal to the incoming repeat sequence of
MDS2)}
\label{intro}
\end{center}
\end{figure}

The process of decoding {\it individual} gene structures is
therefore what interests us here. In the simplest case, micronuclear
versions of macronuclear genes contain many short, noncoding
sequences called {\it internal eliminated sequences}, or IESs. These
are short, AT-rich sequences, and, as their name suggests, they are
removed from genes and destroyed during gene assembly. They separate
the micronuclear version of a gene into {\it macronuclear destined
sequences}, or MDSs (Fig.~\ref{intro}a). When IESs are removed, the
MDSs making up a gene are ``glued" together to form the functional
macronuclear sequence. In the simplest case, IESs are bordered on
either side by pairs of identical repeat sequences (pointers) in the
ends of the adjacent MDSs (Fig.~\ref{intro}b).

\subsection{Scrambled Genes}

In some organisms, the gene assembly problem is complicated by the
``scrambling" of MDSs within a particular gene. In this situation,
the correct arrangement of MDSs in a macronuclear gene is present in
a permuted form in the micronuclear DNA. For example, the actin I
gene in {\it Oxytricha nova} is made up of 9 MDSs and 8 IESs, the
MDSs being present in the micronucleus in the order
3--4--6--5--7--9--2--1--8, with MDS2 being inverted
\cite{pres-gres-92}. During the development of the macronucleus, the
MDSs making up this gene are rearranged into the correct order at
the {\it same} time as IES excision. Scrambling is often {\it
further} complicated by the fact that some MDSs may be {\it
inverted} (a 180$^{\circ}$ point rotation).

\subsection{Fundamental Questions}

Ciliates are remarkably successful organisms. The range of DNA
manipulation and reorganization operations they perform has clearly
been acquired during millennia of evolution. However, some
fundamental questions remain: what are the underlying molecular
mechanisms of gene reconstruction and how did they evolve, and how
do ciliates ``know" which sequences to remove and which to keep?

Concerning the first question, Prescott proposes \cite{pres98} that
the ``compression" of a working nucleus from a larger predecessor is
part of a strategy to produce a ``streamlined" nucleus in which
``every sequence counts" (i.e., useless DNA is not present). This
efficiency may be further enhanced by the dispersal of genes into
individual molecules, rather than having them being joined into
chromosomes. However, so far we still know very little about the
details and evolutionary origins of this intricate underlying
molecular ``machinery."

We may, perhaps, have more success in attempting to answer the
second question: how are genes successfully reassembled from an
encoded version? In the rest of this section we address this
question from a computational perspective, and describe 
a computational model of the rearrangement process.

The model proposed by Prescott, Ehrenfeucht and Rozenberg
(see, for example, \cite{peg01}), is based on three intramolecular
operations (that
is, a single molecule folds on itself and swaps part of its sequence through
recombination). The actual mechanics of cutting and splicing the DNA
sequences are still not understood, but ciliates clearly contain the 
enzymatics tools (e.g., nucleases, ligases, etc.) needed to perform these tasks.

\begin{figure}
\begin{center}
\includegraphics[width=2.5in]{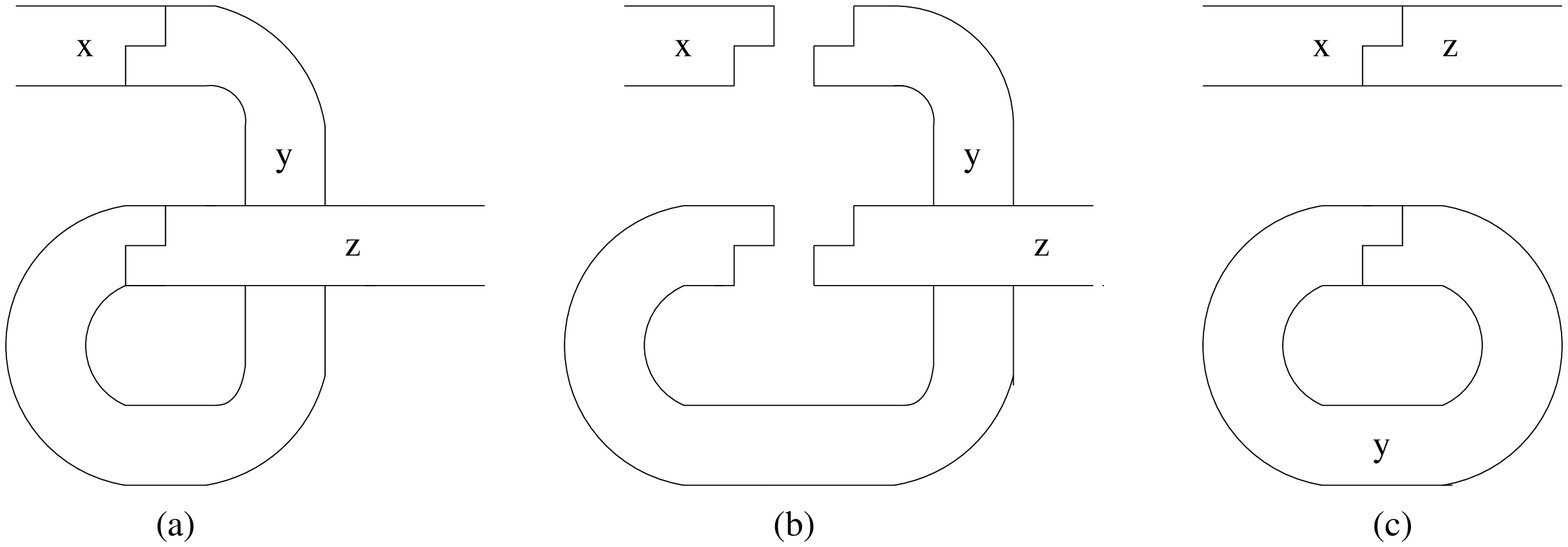}
\caption{Excision} \label{excision}
\end{center}
\end{figure}

The first operation is the simplest, and is referred to as {\it
loop, direct-repeat excision}. This operation deals with the
situation depicted in Fig.~\ref{excision}, where two MDSs ($x$ and
$z$) in the correct (i.e., unscrambled) order are separated by an
IES, $y$.

The operation proceeds as follows. The strand is folded into a loop
with the two identical pointers aligned (Fig.~\ref{excision}a), and
then staggered cuts are made (Fig.~\ref{excision}b). The pointers
connecting the MDSs then join them together, while the IES
self-anneals to yield a circular molecule (Fig.~\ref{excision}c).

\begin{figure}
\begin{center}
\includegraphics[width=2.5in]{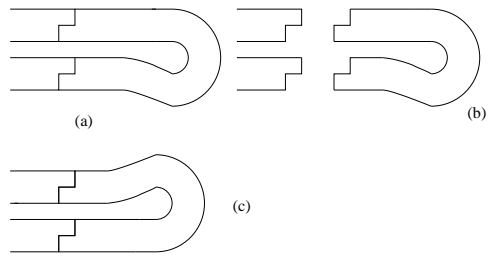}
\caption{Inversion} \label{inversion}
\end{center}
\end{figure}

The second operation is known as {\it hairpin, inverted repeat
excision}, and is used in the situation where a pointer has two
occurrences, one of which is inverted. The molecule folds into a
hairpin structure (Fig.~\ref{inversion}a) with the pointer and its
inversion aligned, cuts are made (Fig.~\ref{inversion}b) and the
inverted sequence is reinserted (Fig.~\ref{inversion}c), yielding a
single molecule.

\begin{figure}
\begin{center}
\includegraphics[width=2.0in]{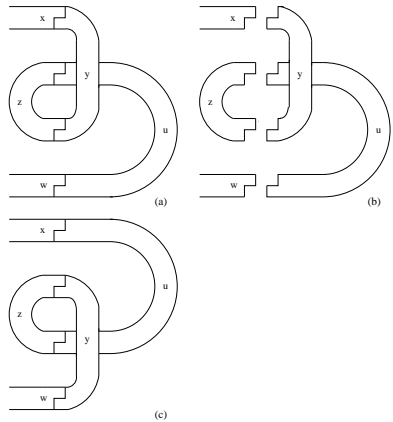}
\caption{Excision/inversion} \label{dlad}
\end{center}
\end{figure}

\noindent The third and final operation is {\it double-loop,
alternating direct repeat excision/reinsertion}. This operation is
applicable in situations where two repeats of two pointers have
interleaving occurrences on the same strand.  The double loop
folding is made such that the two pairs of identical pointer
occurrences are aligned (Fig.~\ref{dlad}a), cuts are made
(Fig.~\ref{dlad}b) and the recombination takes place, yielding the
molecule from Fig.~\ref{dlad}c.

The model has been successfully applied to all known
experimental data on the assembly of real genes, including the actin
I gene of {\it Urostyla grandis} and {\it Engelmanniella mobilis},
the gene encoding $\alpha$ telomere binding protein in several
stichotrich species, and assembly of the gene encoding DNA
polymerase $\alpha$ in {\it Sterkiella nova}. Descriptions of these
applications are presented in \cite{RPOUP2002}. From the perspective of
the current article, the key aspect of this work is that the problem
first originated in the study of a biological system. By expressing the
operation
of NGE in ciliates in terms of an abstract (but biologically plausible)
topological operations,
mathematicians were able to produce a computational model of the process that
appears to account for every decrypted ciliate gene that has been observed
to date. The feedback cycle is then complete when this abstract model
is studied as a novel computational paradigm {\it in its own right} 
\cite{ehpdr04}. Biology and computer science are therefore inextricably
tied together, as neither the computational model nor the descriptions it
offers would be possible without tight interaction between the two disciplines.

% As James Shapiro observes, ``Just as the genome
%has come to be seen as a highly sophisticated information storage system,
%its evolution has become a matter of highly sophisticated information
%processing'' \cite{shapiro99}. If we are to have any hope of understanding
%the workings of NGE, contributions are required from both biologists and 
%computer scientists. The potential payoff -- a new understanding of how NGE
%generates promote novelty in evolution -- may have an impact far beyond 
%either discipline.

\section{Conclusions}\label{section:conclusions}

The overriding message of this paper is that we feel that the bio-inspired computing in general, has reached an impasse.  Through exploring two paradigms (Genetic Programming and Artificial Immune Systems), we have seen that, whilst some significant inroads have been made in the development of systems that in some way, mimic their natural counterpart, there still remains a wide gulf between that the artificial systems can do, compared with the natural systems.  In both cases, we explored the limitations of each approach (a common theme being one of scaling), and concluded that it may be necessary, maybe essential, for each of those paradigms to revisit their biological roots, and take a look from whence they came.  We then considered the areas of evolvable and evolved hardware systems.  These systems, make use of techniques such as GP and AIS to evolve designs and configurations that can be placed into hardware systems, thus bringing benefits such as speed up and so on.  We can conclude from this discussion that whilst it may seem appealing to place solutions into hardware (and in some cases necessary), again we meet the same problem as before, that of scale.  Finally, we reviewed the area of natural genetic engineering.  Here, we showed how close interaction between biologists and
computer scientists has generated a useful model of natural genetic
engineering. This incredibly powerful framework for genomic rearrangement
is one possible explanation of how organisms might  ``confront the issue of
encoding infinity'' \cite{shapiro-transcript}.  
We would say that the other areas of bio-inspired computing may take a lesson from the natural genetic engineering community, and allow us to move forward to create the {\em second generation} of biocomputing systems.

\bibliography{../../../../allbibs}

\end{document}